\pdfoutput=1
\documentclass[11pt]{article}

\usepackage[]{EMNLP2022}

\usepackage{times}
\usepackage{latexsym}
\usepackage{booktabs}
\usepackage{float}
\usepackage{graphicx}
\usepackage{multirow}
\usepackage{arydshln}
\usepackage{graphicx} 
\usepackage{amsmath,amssymb}
\usepackage{makecell}
\usepackage{xcolor}
\usepackage{subcaption}
\usepackage{amsmath,amsfonts,amssymb,amsthm}
\newcommand{\stitle}[1]{\vspace{0.3ex} \noindent{\bf #1}}
\usepackage[clock]{ifsym}
\usepackage[T1]{fontenc}
\usepackage[utf8]{inputenc}

\usepackage{microtype}
\usepackage{enumitem}
\usepackage{inconsolata}

\title{Self-supervised Rewiring of Pre-trained Speech Encoders: \\Towards Faster Fine-tuning with Less Labels in Speech Processing}
\author{Hao Yang\Thanks{~~These authors contributed equally to this work.}$^*$\ \ \ \ \ \ \ \ \ Jinming Zhao$^*$\ \ \ \ \ \ \ \ \ Gholamreza Haffari\ \ \ \ \ \ \ \ \ Ehsan Shareghi \\ \ \ \
Department of Data Science \& AI, Monash University \\
\texttt{firstname.lastname@monash.edu}}

\begin{document}
\maketitle
\begin{abstract}
Pre-trained speech encoders have facilitated great success across various speech processing tasks. However, fine-tuning these encoders for downstream tasks require sufficiently large training data to converge or to achieve state-of-the-art. In text domain this has been partly attributed to sub-optimality of the representation space in pre-trained Transformers. In this work, we take a sober look into pre-trained speech encoders and rewire their representation space without requiring any task-specific labels. Our method utilises neutrally synthesised version of audio inputs along with frame masking to construct positive pairs for contrastive self-supervised learning. When it is used for augmenting the \textsc{wav2vec 2} encoder, we observe consistent improvement of isotropy in the representation space. Our experiments on 6 speech processing tasks, exhibit a significant convergence speedup during task fine-tuning as well as consistent task improvement, specially in low-resource settings.\footnote{Our code and models are available at \url{https://github.com/YangHao97/rewireW2V2}.}

% Ehsan: I will be presenting Mixed as our main configuration and the results for TWIN and Neutral will be presented in the spirit of ablation.

% Recently, pretrained speech models have revolutionized representation learning for speech. However, it is 
% Self-supervised learning approaches have been successfully applied 
%In this work, we aim to improve the expressiveness of these representations. 
% so that they become more data and resource efficient to use.
%Our biggest improvement is to reduce the number of updates from 20 hours to 20 minutes.
% ehsan: need to repalce training speed with convergence speed.
% Pretrained speech transformers have revolutionized representation learning for speech. However,  ... . We propose a novel self-supervised fine-tuning learning approach to rewire the representations of an underlying pretrained speech encoder.  Our lightweight approach (i.e., requiring only 5-6k samples) only utilises neutralized version of audio inputs for forming positive pairs in contrastive fine-tuning and does not rely on any task-specific supervision. When integrated with the widely used wav2vec 2 encoder, our approach  is very effective specifically in low-resource setting, outperforming the wav2vec 2 by a large margin (i.e., up to ...) across 7 speech understanding tasks.

\end{abstract}

\section{Introduction}
Self-supervised pre-trained speech encoders~\cite{hsu2021hubert,baevski2020wav2vec} are universal models that are beneficial to a wide range of speech processing tasks and domains~\cite{liu2022audio,tsai2022superb}. 
%Depending on the pretext tasks, different models have been developed and can be categorized into: generative methods \cite{ravanelli2020multi,ling2020decoar}, predictive methods \cite{chen2021wavlm,hsu2021hubert} and contrastive methods \cite{baevski2020wav2vec,chung2021w2v}. 
Similar to other modalities such as text, these pre-trained encoders are fine-tuned towards downstream tasks~\cite{wang2022wav2vec,gallego2021end}. While the fine-tuning step often benefits substantially from the presence of warm pre-trained data encoders, for involved tasks such as Automatic Speech Recognition (ASR), it still requires both sufficiently large training sets and several iterations~\cite{DBLP:conf/interspeech/YangCCLLLLSCLHT21} for convergence to an acceptable task performance.  

Side-stepping the size of the parameter space as a well-studied challenge for fine-tuning Transformer models, a confounding factor contributing to this issue, which has been recently discussed for text domain~\cite{su-etal-2022-tacl, gao2021simcse, liu2021fast, su2021whitening}, is the sub-optimal utilisation of the representation space (e.g., anisotropy~\cite{ethayarajh-2019-contextual}). This is of paramount importance since speech, unlike text, carries information (e.g., prosodic and para-linguistic) beyond content which demands a richer utilisation of the representation space~\cite{mohamed2022self}. Inevitably, less expressive initial representations translate into longer training and call for more labelled data, even in cases of frozen models. Nonetheless, understanding representation space utilisation in pre-trained speech Transformers is heavily under-explored~\cite{pasad2021layer,hsu2021robust}.

We move towards addressing this gap by highlighting the properties of such representation spaces, and proposing a self-supervised learning method that improves their utilisation prior to task fine-tuning. 
% The goal is to rewire the geometric property (i.e., injecting isotoropy) of the existing representation space instead of injecting new knowledge to it.
%\footnote{In the presence of explicit supervision signal, any desired property could be injected during representation learning, we focus only on pre-training in self-supervised setting.} 
Our contrastive learning framework constructs positive pairs by (i) encouraging invariance to local perturbations both at the input 
%level via frame masking and truncation, and at the 
and representation levels, and (ii) enhancing sensitivity to content by using monotonically synthesised version of speech inputs. 
%pre-trained speech transformers (i.e., \textsc{wav2vec~2~large}(\textsc{w2v2})~\cite{baevski2020wav2vec}) 

% \textcolor{red}{To ensure the results cover diverse aspects of speech, including content, speaker and semantics, }
Our experimental findings across 6 diverse speech processing tasks (covering content, speaker and semantics tasks), built on top of the widely used \textsc{wav2vec~2~large} (\textsc{w2v2})~\cite{baevski2020wav2vec} encoder, demonstrate that contrastive rewiring brings substantial improvement, both in task performance and fine-tuning speed. Particularly, our approach shines in the low-resource condition, outperforming the \textsc{w2v2} baseline with substantially fewer number of fine-tuning updates. For instance, in ASR with 1\% training data,  our approach achieves $1/4$ of the error in $1/5$ of fine-tuning updates. Beyond task performance and convergence speed, both our qualitative and quantitative  analyses on the representation space highlight the improvements injected by our rewiring strategy.

\section{Self-Supervised Contrastive Rewiring}
%\ehsan{1p: an overview of what we are trying to achieve and key benefits of our method needs to be added.} 
Our method builds on top of a pre-trained speech encoder, by using a small (less than $7k$) set of raw unlabelled audio signals to form the self-supervised learning basis for contrastive rewiring. In what follows, we detail how utterance-level speech representations are produced from the underlying encoder, and provide a brief overview of the InfoNCE objective function used for our contrastive rewiring. We finish by explaining how we construct the pairs needed for contrastive learning. 
%Here, we detail how speech representations are generated from \textsc{w2v2} (\S\ref{sec:rep}), then we describe how positive pairs are constructed (\S\ref{sec:pos}), and we finish b first and then describe our training objective. 
% In this section, we will detail how positive pairs are constructed first and then describe our training objective.  

% The formation of positive and negative samples is at the core of contrastive learning. While negative pairs (i.e., ignoring their difficulty) are typically obtained from randomly sampled data within a batch, construction of useful positive pairs is more challenging. In this section, we will detail how positive pairs are constructed first and then describe our training objective.  
\paragraph{Speech Representation.} Most pre-trained speech encoders, including \textsc{w2v2}, do not have an explicit token representing utterance-level representation~(e.g., [CLS] for BERT~\cite{kenton2019bert}). Given a raw audio sequence $s$ of length $L$, \textsc{w2v2} emits $m$ vectors, %$\mathbf{H}=\{\mathbf{h_1},\mathbf{h_2},...,\mathbf{h_m}\}$ 
where $m \ll L$, at each layer (total of 24 Transformer layers + 1 feature extractor layer). Similar to \citet{chung2021similarity}, we take the mean of these vectors to construct the utterance-level representation used for contrastive learning. 

\paragraph{InfoNCE.} We use the InfoNCE objective \cite{oord2018representation} to rewire speech representations by pulling positive examples, $(s_i,s'_i)$, closer and pushing away the negative pairs, $(s_i,s_j)$. The loss for a batch $b$ of size $|\mathcal{D}_b|$ is,
% \scalebox{0.8}{ 

{\footnotesize
\begin{equation*}
\mathcal{L} = -\sum_{i=1}^{ |\mathcal{D}_b|}\log\frac{\exp(\texttt{cos}(f(s_i), f(s'_i))/\tau)}{\displaystyle \sum_{s_j\in N_i \cup \{s'_i\}}\exp(\texttt{cos}(f(s_i), f(s_j))/\tau)},
\end{equation*}}%
% }%
where $f(.)$ indicates the encoder, $\tau$ denotes the temperature hyperparameter, $\texttt{cos}(.,.)$ denotes the cosine similarity between two representations, $N_i$ includes all negative examples for $s_i$. All parameters of the encoder are updated during optimisation.
%of the loss.
\begin{figure}[t]
  \centering
    \includegraphics[width=1\linewidth]{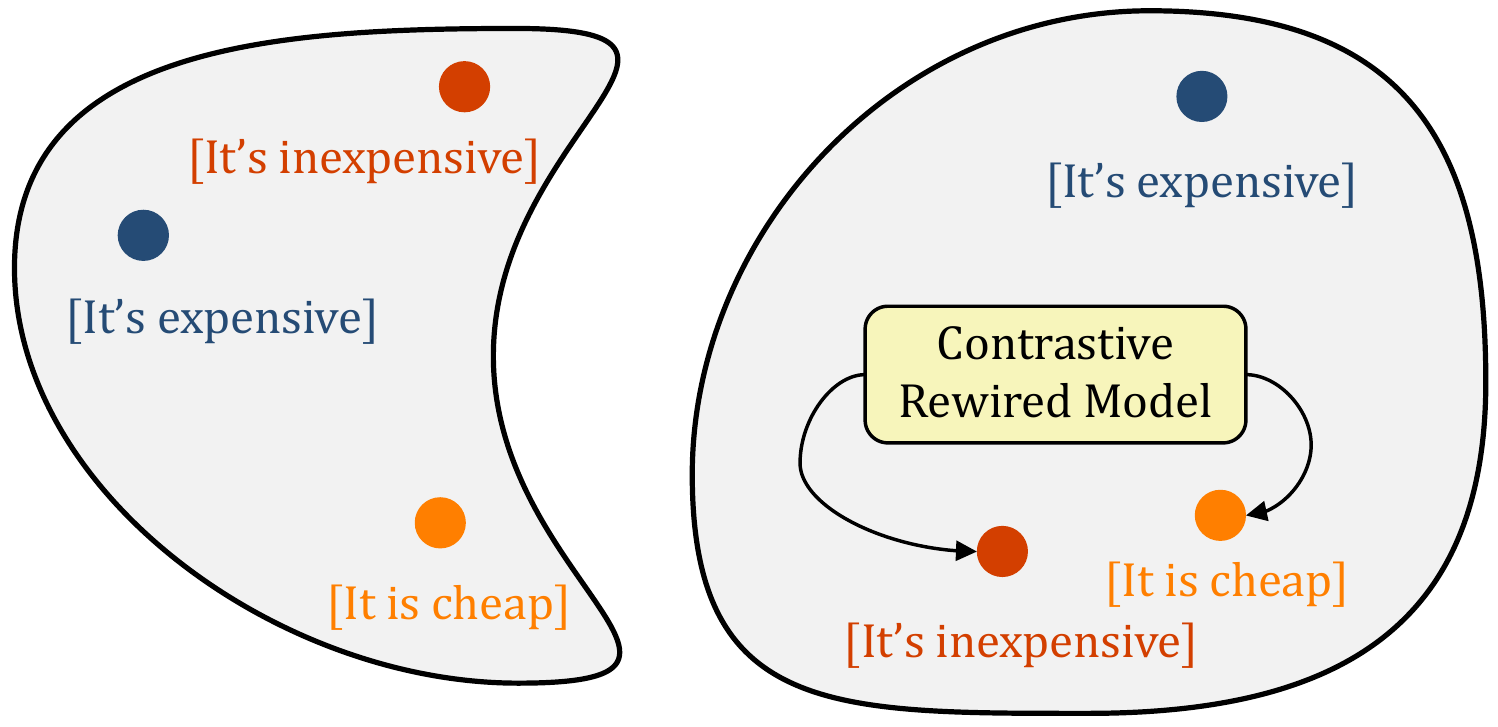}
 \caption{Conceptual visualisation: Vanilla representation space which is very sensitive to surface similarity of audio signals (left) vs. rewired  representation space with \texttt{Neutral} strategy which places more emphasis on content similarity (right).}
  \label{fig:arch}
\end{figure}
\subsection{Contrastive Pair Construction}
\stitle{Positive Pairs.} We form positive pairs both at the raw and representation levels. For a given audio signal $s_i$, we deploy the following 3 strategies to construct  its corresponding positive pairs, ($s_i,s'_i$):
\begin{description}[leftmargin=2.5mm,parsep=0pt,partopsep=0pt]
\item[\texttt{Twin}.] Inspired by \citet{liu2021fast,gao2021simcse}, given a speech sequence of length $L$, we first duplicate it. Then we randomly select a starting point for a span, and mask $p\times L$ consecutive signals from the audio, replacing them with $\texttt{[MASK]}$. We use $p=20\%$ in our experiments. This is applied always and only once to each $s_i$. 
%between 0 and $(1-n\%)\times L$ and replace consecutive $n\%\times L$
%We expect that the missing parts of speech information can be reconstructed by contrastive learning models.  
\item[\texttt{Neutral}.] For a given audio $s_i$, its monotonic neutral version is created from available transcripts\footnote{Alternatively, one can apply an off-the-shelf ASR first over speech to produce transcripts when transcripts are absent.} using Festival Speech Synthesis System.\footnote{\url{http://festvox.org/festival}} The synthesizer is chosen because it is able to produce non-expressive speech, as demonstrated in previous studies~\cite{lotfian2017building}.   
% \textcolor{red}{(ref) proposed the speech from the festival synthesizer is considered as neutral speech, or not other TTS systems.} 
The neutral version is devoid of noise, prosody and para-linguistic features, focusing mostly on content. Figure~\ref{fig:arch} illustrates a visualisation of the desired expected effect from \texttt{Neutral} rewiring.
%It is chosen as alternative positive speech signals, as many speech understanding
%\footnote{We exclude generation tasks as they are more difficult than classification tasks and more is involved (e.g., decoding strategy).} 
%tasks are semantic tasks which require linguistic information \cite{DBLP:conf/ijcai/QinXC021}. 
%The hope is that content information can be better highlighted or disentangled in the feature space with our proposed method. 
\item[\texttt{Mixed}.] While the \texttt{Twin} strategy aims to make the representations invariant to local changes and noise, the \texttt{Neutral} approach tends to rewire the space based on content-level similarity. To leverage the benefits of both worlds, as our main strategy, we uniformly interchange \texttt{Twin} and \texttt{Neutral} in the \texttt{Mixed} setting. 
\end{description}

\stitle{Negative Pairs.} In all strategies, given a batch $b$ and a sample $s_i~\in~b$, the set $N_i$ of negative examples for $s_i$ is $N_i=\{s_j|s_j\in b,j\neq i\}$. Further, we have specific negative samples added to $N_i$ per each strategy to construct negative pairs, $(s_i,s_j)$: 
\begin{description}[leftmargin=2.5mm,parsep=0pt,partopsep=0pt]
\item[\texttt{Twin}.] $N_i\cup \{\texttt{twin}(s_j)|s_j\in b,j\neq i\}$.
\item[\texttt{Neutral}.] $N_i\cup \{\texttt{neutral}(s_j)|s_j\in b,j\neq i\}$.
\item[\texttt{Mixed}.] Union of the above two.
\end{description}
Similar to \citet{liu2021fast} and \citet{gao2021simcse}, in all our strategies, we apply dropout to perturb both the representations and internal components of Transformer. Note that \texttt{Twin} and masking are aligned with how a Transformer-based model is trained, so rewiring with this strategy is unlikely to create conflicts. Furthermore, \texttt{Neutral} potentially pushes representations towards eliminating paralingual feature from speech representations, as it shifts the focus to content.

\section{Experiments}

In this section we describe our experimental settings (\S\ref{sec:expset}) followed by downstream task results in full and low-resource scenarios (\S\ref{sec:main_results}). We finish by providing an analysis on the quantitative and qualitative properties that were improved via our contrastive rewiring approach (\S\ref{sec:analysis}).
\subsection{Experimental setups}\label{sec:expset}

\stitle{Rewiring Dataset.} For all contrastive rewiring strategies  on top of a pretrained \textsc{w2v2}, we used a small subset (6.6k instances) of LibriSpeech~\cite{panayotov2015librispeech} train-clean-100 where instances lengths are under 180k, which is also part of training data for W2V2.
% \textcolor{red}{the small set of samples are from the same domain as the pretraining data of W2V2.} 
LibriSpeech also contains transcripts, which we used for \texttt{neutral} rewiring to produce neutral speeches.
\begin{table*}[t]
\small
    \centering
    \scalebox{1}{
    \begin{tabular}{l c c c c c c c}
        \toprule
         & & & &\multicolumn{4}{c}{\# Training Instances} \\
         \cmidrule(lr){5-8}
         \thead{Task} & Eval. &\thead{Type} & \thead{Avg.Src Len} &1\%&5\%&10\%&100\%\\
         \cmidrule(lr){1-4}\cmidrule(lr){5-5}\cmidrule(lr){6-6}\cmidrule(lr){7-7}\cmidrule(lr){8-8}
    SD &DER$\downarrow$& speaker  & 233k &  0.14k & 0.70k & 1.39k & 13.9k      \\
    SF &F1$\uparrow$& semantic  & 45.6k & 1.05k & 5.24k & 10.5k & 104.7k \\
    IC &Acc$\uparrow$&  semantic & 36.6k & 0.23k & 1.16k & 2.32k & 23.1k \\
    KS &Acc$\uparrow$&  content & 15.7k & 0.51k & 2.56k & 5.11k & 51.1k\\
    % QbE & content  & 107.5k & 2438 || 138 || 138      \\
    ASR &WER$\downarrow$& content  & 201k & 0.28k & 1.43k & 2.86k & 28.5k       \\
    QbE &MTWV$\uparrow$& content  & 107k & $*$ & $*$ & $*$ &     $*$\\
    \bottomrule
    \end{tabular}}
    \caption{Dataset statistics. $*$: QbE is zero-shot. Tasks types are determined by SUPERB.}
    % \vspace{-3mm}
    \label{tab:data}
\end{table*}

\stitle{Downstream tasks.} We experimented with 6 diverse downstream speech processing tasks from SUPERB benchmark~\cite{DBLP:conf/interspeech/YangCCLLLLSCLHT21}: Automatic Speech Recognition (ASR), Speaker Diarization (SD), Intent Classification (IC), Slot Filling (SF), Keyword Spotting (KS) and Query by Example Spoken Term Detection (QbE). These tasks cover semantic, speaker, and content in speech tasks. For each task, we simulate various resource conditions by sampling 1\%, 5\%, 10\% of the entire training set, while using the original dev set to decide the best model. Original test sets were used for evaluation. Each task is evaluated using its specific evaluation metrics. For details on tasks and evaluation metrics, see \emph{Appendix}~\ref{sec:appendix_tasks}. We detail data statistics in Table~\ref{tab:data}. We follow the instructions in SUPERB and use the \texttt{s3prl} toolkit\footnote{https://github.com/s3prl/s3prl} to prepare the datasets. 

% \mich{[Hao: Please modify accordingly. For instance, provide more details on the dev/test for each dataset.]}  

% Ehsan: weird last column ...

    % Our experiments are based on SUPERB pipeline which is a universal performance benchmark and used to evaluate the performance of pre-trained models’ representations. In SUPERB, the pre-trained model only provides speech representations, and the parameters of the pre-trained model do not change when fine-tuning on downstream tasks. We conduct experiments on 7 downstream tasks in SUPERB: Phoneme Recognition(PR) and Automatic Speech Recognition(ASR) adopt LibriSpeech train-clean-100/dev-clean/test-clean subsets, Keyword Spotting(KS) adopts Speech Commands dataset v1.0 , Query by Example Spoken Term Detection(QbE) adopts the English subset in QUESST 2014, Speaker Diarization(SD) adopts LibriMix dataset, Intent Classification(IC) adopts Fluent Speech Commands dataset and Slot Filling(SF) adopts Audio SNIPS. 

%\subsection{Model Settings and Experiment Settings}    
\stitle{Baseline.} While our approach is not dependent on a specific  speech Transformer, we use \textsc{w2v2} as the most widely used  Transformer-based speech encoders. We follow the SUPERB evaluation pipeline by freezing it as an encoder for downstream tasks, while attaching a  benchmark-specified lightweight prediction head for each task. For details on task head architectures, see \emph{Appendix}~\ref{sec:appendix_tasks}. We follow identical protocol for fine-tuning and evaluating all models.

%\paragraph{Ours} During the rewiring process, we fine-tune \textsc{wav2vec 2} with the LibriSpeech subset, while setting the maximum audio length to 18k to avoid memory issues. Then for downstream tasks, same as the settings for the baseline, we keep the encoder frozen and attach heads to these tasks. This applies to all \textsc{Twin}, \textsc{Neutral} and \textsc{Mixed}.
%To mask the augmented data for training \textsc{Twin}, we randomly select a point in the first four-fifths of the signals and mask consecutive signals of one-fifth of its total length. For \textsc{Neutral}, we generated neutral speech from the transcriptions of the chosen LibriSpeech subset offline, with the aforementioned TTS software. For \textsc{Mixed}, we randomly select a technique from \textit{twin} and \textit{neutral} to construct positive pairs. 
% \subsection{}
\stitle{Implementation details.} 
During rewiring, we set the dropout to 0.1, the learning rate to 1e-6 and the temperature for InfoNCE to 0.04. To overcome the hardware constraint, we truncated audio signals and encoded them sequentially. To avoid memory issues that come from lengthy audio signals, we read one audio data at a time and set the audio length threshold to 90k. Our batch size for rewiring was 8, and we rewired \textsc{w2v2} for 1.7k, 11.6k, 5k updates for \texttt{Twin}, \texttt{Neutral} and \texttt{Mixed}, respectively.

%2438$*$
During downstream task fine-tuning, under $100\%$ condition, we set the max step to $\sim 200k$ for both the baseline and our models.  Under the 1\%, 5\% and 10\% resource settings, we fine-tune the models for substantially less number of updates, although the baseline model still requires several steps for convergence (See Table~\ref{table:main}). We follow the settings in \citet{DBLP:conf/interspeech/YangCCLLLLSCLHT21} for hyper-parameters, and use weighted-sum of hidden states of each layer from \textsc{w2v2} as the final representation for downstream tasks. For more details, please refer to \emph{Appendix}~\ref{appendix:training details}. 
\begin{table*}[t]
\setlength{\tabcolsep}{7pt} %2.5pt
\centering
\scalebox{0.7}{ %0.6
\begin{tabular}{lccccccccccc}
\toprule
% &&\multicolumn{7}{c}{\bf Tasks}\\
% \cmidrule(lr){3-9}
%  Tr.&Model& SF  & SD  & QbE & KS & IC & PR$^*$ &ASR$^*$ \\
% &&\multicolumn{14}{c}{\bf Tasks}\\
 & & \multicolumn{2}{c}{SD}  & \multicolumn{2}{c}{SF}  &  \multicolumn{2}{c}{IC} & \multicolumn{2}{c}{KS} & \multicolumn{2}{c}{ASR} \\
 \cmidrule(lr){3-4}\cmidrule(lr){5-6}\cmidrule(lr){7-8}\cmidrule(lr){9-10}\cmidrule(lr){11-12}
 Tr.&Model& DER $\downarrow$  &\VarClock& F1 $\uparrow$ &\VarClock  & Acc $\uparrow$ &\VarClock & Acc $\uparrow$ &\VarClock &  WER $\downarrow$&\VarClock\\
 \midrule 
 \bf \parbox[t]{2mm}{\multirow{4}{*}{\rotatebox[origin=c]{90}{$1\%$}}} 
 &\textsc{W2V2}&10.23 & 6.7k &57.88 & 42k& 12.54 & 6.8k & 85.17&9.2k &  99.99 & 50k\\
 &\texttt{Twin}&  \underline{9.04} &\textbf{0.1k}&\underline{65.14} &38k & 7.03 &  \textbf{0.2k} &85.98& 2k& \underline{23.43}& 12k\\
 &\texttt{Neutral}&  12.73 &0.4k &63.05 &36k &  35.48 &0.4k&\underline{93.08}& \textbf{0.25k} & 36.20 & 12k\\
 &\texttt{Mixed}&12.30 &0.2k& 64.04 &\textbf{17k}& \underline{43.05} &0.8k& 90.58&1.3k& 26.35&\textbf{10k}\\
  \midrule
   \bf \parbox[t]{2mm}{\multirow{4}{*}{\rotatebox[origin=c]{90}{$5\%$}}} 
 &\textsc{W2V2}&9.20 &4k  &78.29 &58k& 53.07 &16k&94.25&20k & 14.70 & 100k\\
 &\texttt{Twin}& \underline{8.15} &0.4k&\underline{82.65} &\textbf{56k}& 39.41 &\textbf{2.8k}&94.12&3.75k& \underline{8.77} & \textbf{46k}\\
 &\texttt{Neutral}& 10.01 &1.2k & 79.48 & 62k& \underline{91.27} &4.3k &94.28&1.75k& 21.12& 40k\\
 &\texttt{Mixed}&10.39 &\textbf{0.2k}& 81.00 &68k & 90.29 &  3.1k &\underline{94.77}&\textbf{1.25k}& 14.25& 48k\\
  \midrule
   \bf \parbox[t]{2mm}{\multirow{4}{*}{\rotatebox[origin=c]{90}{$10\%$}}} 
 &\textsc{W2V2}&8.21 &6k &80.74 & 90k& 77.91 &45k & \underline{95.85}&15.5k& \underline{5.96} &\textbf{90k}\\
 &\texttt{Twin}& \underline{7.70} &1.6k&\underline{85.00} & 88k&57.97 &\textbf{1.6k}& 94.97&4.5k& 6.45& 92k\\
 &\texttt{Neutral}&  8.57 & 2.4k &82.30 & 84k& 92.75 &15k&  94.90&4.5k & 16.72& 98k\\
 &\texttt{Mixed}&9.60 &\textbf{0.6k} &84.02 &\textbf{74k} &\underline{93.67} &2k &95.07&\textbf{1.5k} & 11.27& 93k\\
%  & $\triangle$ \\
  \midrule
   \bf \parbox[t]{2mm}{\multirow{4}{*}{\rotatebox[origin=c]{90}{$100\%$}}} 
 &\textsc{W2V2$^\diamond$}&\underline{5.68} & 27.6k &87.67 &150k&94.60 &90k& 96.64& 55k& \underline{3.78}$^*$ &166k\\
 &\texttt{Twin}&5.95 &6.5k&\underline{89.93} & 160k&  93.36 &55k&96.62&\textbf{15k} &4.07$^*$ &\textbf{80k}\\
 &\texttt{Neutral}&6.67 & 11k & 88.05 &\textbf{125k} &\underline{97.18} &15k& 96.30 & 20k &10.34$^*$ &105k\\
 &\texttt{Mixed}&6.73 &\textbf{4.5k}&89.18 & 140k& \underline{97.18} &\textbf{5k}& \underline{96.75}&25k &7.46$^*$ &110k\\
%  & $\triangle$ \\
 \bottomrule
 \multicolumn{12}{c}{QbE (100\%) : [\textsc{w2v2$^\diamond$}:4.93], [\texttt{Twin}:5.04], [\texttt{Neutral}:3e-10], [\texttt{Mixed}: \textbf{7.50}]}\\ 
 \bottomrule
\end{tabular}
}
\caption{Results and the number of fine-tuning updates to achieve the best performance for downstream tasks in various resource conditions. $^*$: for ASR the number of training instances was $76\%$, as remaining overlapped with the data used for rewiring. $^\diamond$: replicated.}
\label{table:main}
% \vspace{-2mm}
\end{table*}
% \mich{$\triangle$: refers the average improvement of our three methods.} 
\subsection{Main Results}\label{sec:main_results}
Our task fine-tuning results are presented in Table~\ref{table:main}. The results highlight the improvement our contrastive rewiring brings across various downstream tasks, both in performance and fine-tuning speed. Notably, in 1\% and 5\% training scenarios, our models (at least 1 and in many cases all 3 strategies) outperform the \textsc{w2v2} baseline on all tasks, while requiring substantially less number of fine-tuning updates. More details per task follows:
\begin{description}[leftmargin=2.5mm,parsep=0pt,partopsep=0pt]
\item[\texttt{IC}] The benefit of rewiring in the low-resource~(1\%) condition is massive. Even at 100\%, our method outperforms \textsc{W2V2} without rewiring. The best rewiring setting helps to speedup convergence by an average of 22$\times$ in various data conditions. 
\item[\texttt{SF}] Similarly, rewiring is beneficial in all settings, even at 100\%. As the training data size decreases, the benefit of rewiring increases.
\item[\texttt{SD}] Our method is helpful in low- and mid-resource settings; at 100\% it has lower performance, but \texttt{Mixed} helps the model to converge 6$\times$ faster.
\item[\texttt{KS}] Rewiring boosts performance and convergence significantly at 1\%. It has the similar performance in other conditions, but still helps training with an average speedup of 10$\times$.
\item[\texttt{ASR}] Our \texttt{Mixed} model at 1\% achieves an error rate of 26 after 10k updates, whereas \textsc{w2v2} has an error rate of 99.99 even after 50k updates. The gain from rewiring is also noticeable at 5\%, while levelling out afterwards.
\item[\texttt{QbE}] Rewiring improves performance significantly with our \texttt{Mixed} model yielding 7.50 in MTWV over 4.93 produced by the vanilla \textsc{W2V2}.
% brings a performance gain of 44\% at this zero-shot task.
\end{description}
Overall, our proposed approach improves performance and convergence across various speech processing tasks in all resource conditions, with more remarkable gains in low- to mid-resource conditions.

\subsection{Analysis and Discussion}\label{sec:analysis}
\begin{figure}[t]
    \centering
    \includegraphics[trim={6cm 3.3cm 4.5cm 3.5cm},clip, scale=0.12]{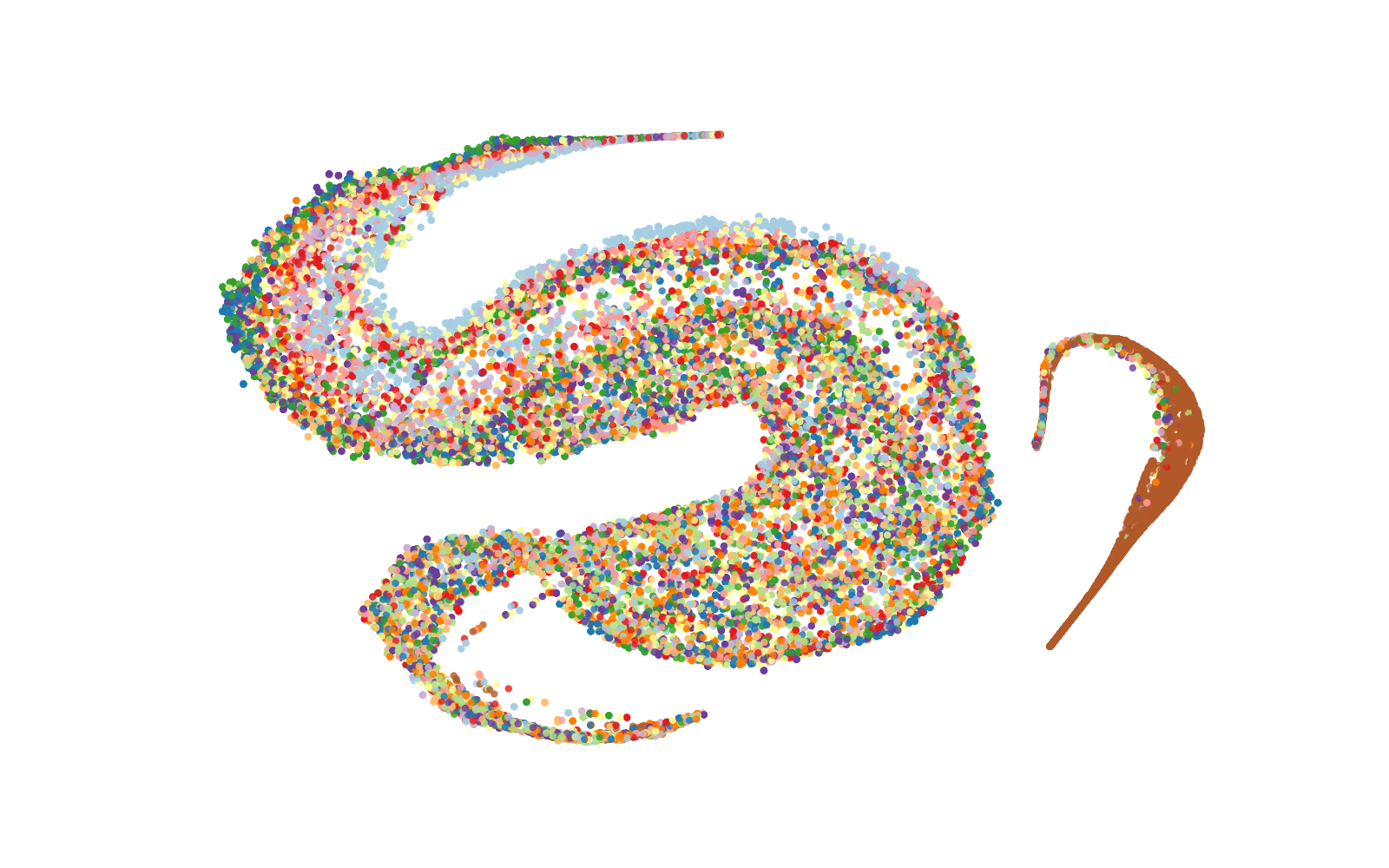}\hspace{3mm}
    \includegraphics[trim={6cm 3.3cm 4.5cm 3.5cm},clip, scale=0.12]{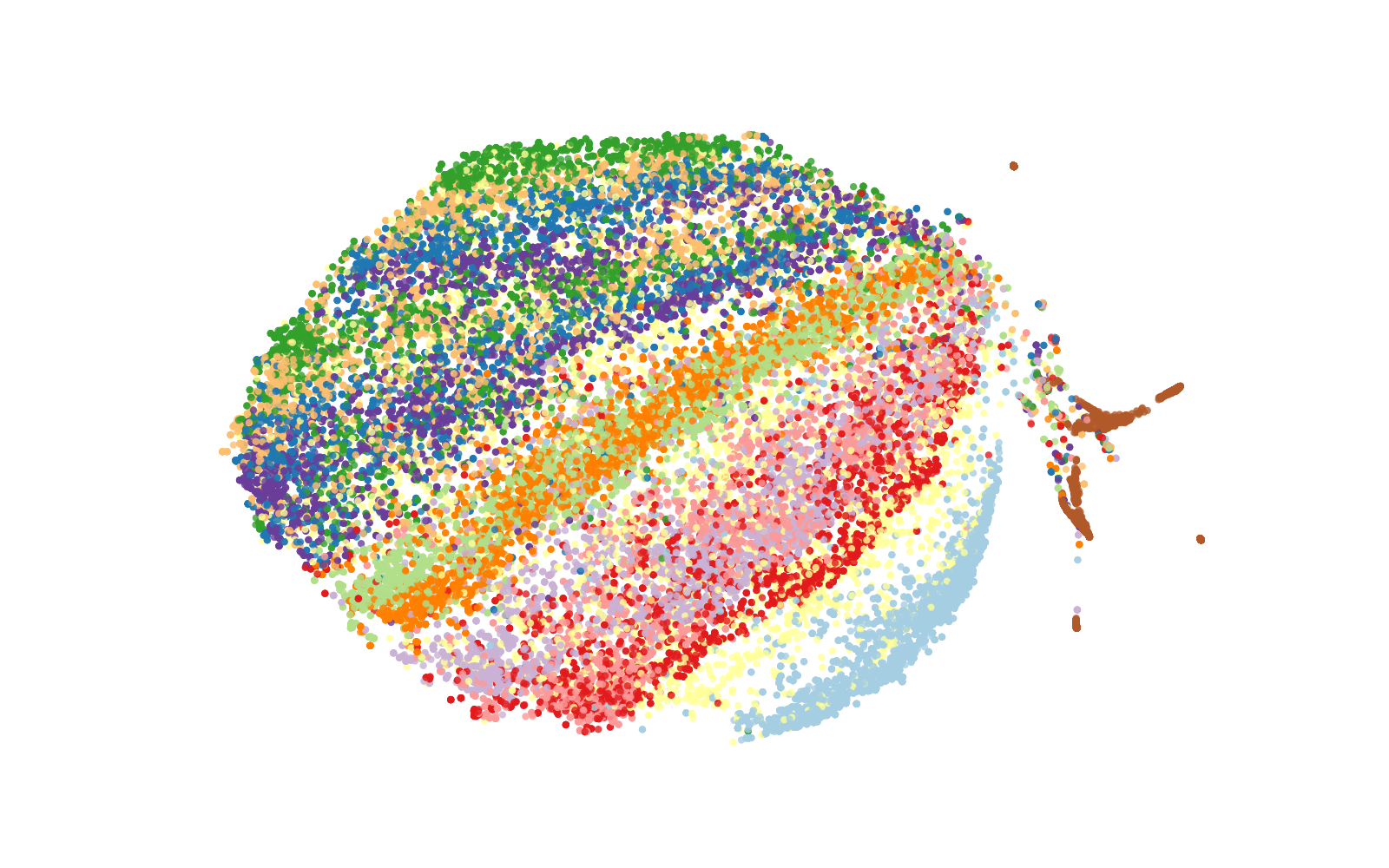}
    \includegraphics[trim={6cm 3.3cm 4.5cm 3.5cm},clip, scale=0.12]{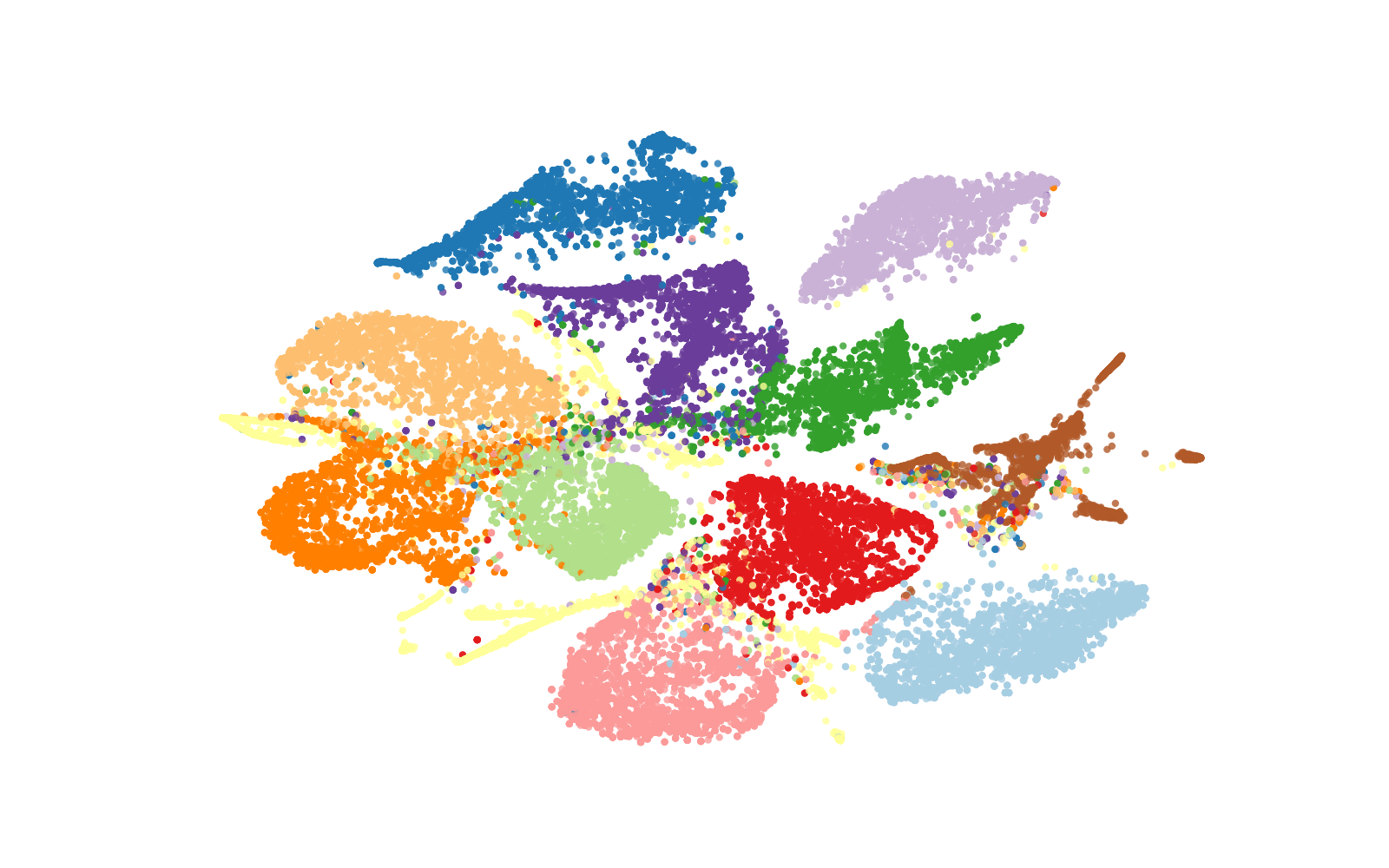}\hspace{3mm}
    \includegraphics[trim={6cm 3.3cm 4.5cm 3.5cm},clip, scale=0.12]{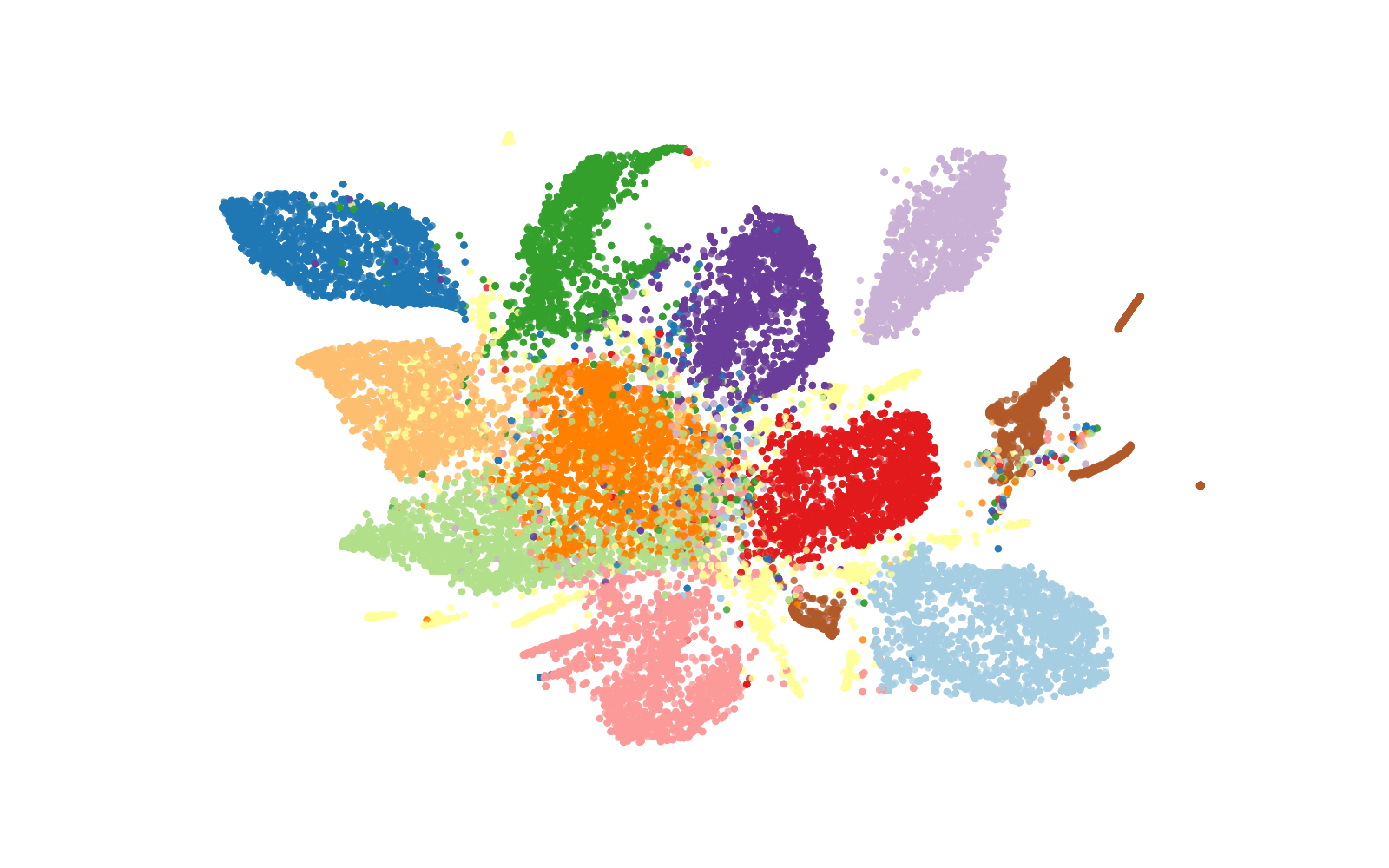}
    \caption{t-SNE visualisations of the representations from Keyword Spotting~\cite{DBLP:conf/interspeech/YangCCLLLLSCLHT21}  training set prior to fine-tuning. \textbf{Top-Left:} \textsc{W2V2}; \textbf{Top-Right:} \texttt{Twin}; \textbf{Bottom-Left:} \texttt{Neutral}; \textbf{Bottom-Right:} \texttt{Mixed}. Colours indicate class labels.}
    \label{fig:tsne}
    % \vspace{0.1cm}
\end{figure}

\stitle{Qualitative Analysis.} Figure~\ref{fig:tsne} demonstrates the {t-SNE}~\cite{JMLR:v9:vandermaaten08a} visualisation of the impact of applying each strategy to \textsc{w2v2}. We clearly observe that better clustering of the representation space, specially after applying $\texttt{neutral}$ (bottom-lsumeft), emerges without any task fine-tuning. 
% \begin{itemize}
%     \item grouping the tasks in terms of their reliance on local/global phenomena (i.e., is content important or not, is it just a global aspect of data that matters for the task, etc) to analyse why contrastive learning works in general or why one strategy works/fails
%     \item Read section 4.4 of MirrorBERT paper to do similar analysis here
%     \item Isotropy
% \end{itemize}

\stitle{Task Type.} According to Table~\ref{tab:data}, we have 3 types of tasks, content, semantic and speaker. As expected the content and semantic tasks are the biggest gainers in the low-resource setting (1-5\%) from our contrastive rewiring. This verifies our earlier qualitative analysis and aligns well with the motivation for leveraging neutral speeches in the \texttt{Neutral} strategy which is expected to put more emphasise on content.

\stitle{Isotropy.}
We speculate that the benefits of our rewiring strategies also roots in reshaping the representation space geometry. The isotropy of the embedding space is a desired property and we conjecture that the representations of \textsc{w2v2} are potentially anisotropic: crowded into a narrow slices of the representation space. To test our conjecture, we calculated the isotropy scores of speech representations produced by 4 models on 5 different datasets. We approximate the isotropy score~\cite{mu2018allbutthetop}, $$IS(\mathcal{V})=\frac{\min_{m\in \mathcal{M}}\sum_{v\in \mathcal{V}} \exp(m^\intercal v)}{\max_{m\in \mathcal{M}}\sum_{v\in \mathcal{V}} \exp(m^\intercal v)},$$ where $\mathcal{V}$ is the matrix of representations, and $\mathcal{M}$ is the set of eigen vectors of $\mathcal{V}^\intercal\mathcal{V}$.
%
%We argue that the benefits of our three models also stem from its reshaping of the embedding space geometry. The isotropy of the embedding space is a favorable property for evaluating the quality of sentence representations. We conjecture that the representations of \textsc{wav2vec 2}, also based on transformer, are just as anisotropic as BERT: they are crowded into a narrow vector space. To test our conjecture, we calculated the isotropy scores of speech representations produced by 4 models on 5 different datasets of 6 downstream tasks.
%
The isotropy scores of \textsc{w2v2}, \texttt{Twin}, \texttt{Neutral}, and \texttt{Mixed} models are on the order of 1e-300, 1e-10, 1e-30, and 1e-10 respectively. This result confirms our conjecture that our three models improve isotropy of the representation space by orders of magnitude compared with \textsc{w2v2}.
%, while still very small compared with text counterparts.

\section{Conclusion and Future Work}
In this paper,
% \textcolor{red}{ to show the sub-optimal utilisation of representation in a currently widely used speech Transformer model, and how we could improve the utilisation of the representation space in such a model by a light rewiring,} 
we presented effective and efficient self-supervised contrastive learning methods to  rewire the representations of speech pre-trained Transformer model. We demonstrated that lightly rewiring \textsc{wav2vec 2} improves the convergence speed of fine-tuning as well as task performance on 6 downstream tasks. In particular, in low-resource condition our method performed substantially better than the underlying \textsc{wav2vec 2}. Our analysis indicated the rewiring has created a much better discriminated representation space, making it better suited for fine-tuning towards tasks. As future work, we plan to cover more downstream tasks, and invest more into designing hard negative pairs to further augment the contrastive learning.

%continue to dig deeper into why our models are only effective on most tasks, but not all tasks.

% \begin{itemize}
%     \item recap the findings
%     \item as future work: forming harder negative pairs by utilising ngram similarity over transcripts
% \end{itemize}
% \clearpage
\section{Limitations}
We hoped to extend our experiments to all tasks on SUPERB but certain tasks involved data access difficulties (we initiated the requests but never got access). Additionally, we did not see significant gain with rewiring on Phoneme Recognition, which could stem from our construction of utterance-level representation. This suggests finer grained granularity of representations need to be included for fine-grained tasks. It also requires further investigation to fully understand why our method performs extremely well in certain source conditions and relatively well on certain tasks compared to others. Although we have provided certain conjectures, this analysis requires a standalone work.\\ 
% \mich{PR doesn't work well.}
% {\color{brown}[Hao: I'm not sure if we should include this conjecture of mine, it's a strong point but I have no direct evidence only circumstantial evidence. We might be able to dig into it later and be our next work. So I'm not sure if we need to publish it.]}Our conjecture is: In text domain, two similar text representations usually mean that the meaning of these two sentences are similar. Because the text representation is only derived from the words in the text sentence. But in speech domain we can't think that way. Two similar speech representations may mean that the waveforms of the two audios are similar, not that the meanings of the two audios are similar. Because the waveform of the audio is not only determined by the content of the audio, but also different speakers, tones, accents, noise, speed of speaking, etc. will also have a very large impact. In this case, the original representation in the vector space is crowded together , after we use the masking technique, the vector space will indeed be expanded, but the vectors in the vector space will form many small clusters. Each small cluster will contain many representations with different meanings but similar waveforms. This means that there may be many representations with different meanings being pulled closer, and some sentences with the same meaning being pushed further. This greatly affects the performance of the model. Figure x shows the results.
\section{Ethics Statement}
Our work is built on top of \textsc{wav2vec 2}, which is pretrained on massive speech data. Our goal was not to attend to alleviate the well-documented issues (e.g., privacy, undesired biases, etc) that large pretrained models have. For this reason, we share the similar potential risks and concerns posed by these models. 
% But our experimental approach does not change the underlying issues with pre-trained models, we aim to demonstrate the usability of rewiring on speech pre-trained models. So we inherently have similar potential risks posed by these models.
% Scientific work published at EMNLP 2022 must comply with the \href{https://www.aclweb.org/portal/content/acl-code-ethics}{ACL Ethics Policy}. We encourage all authors to include an explicit ethics statement on the broader impact of the work, or other ethical considerations after the conclusion but before the references. The ethics statement will not count toward the page limit (8 pages for long, 4 pages for short papers).

\bibliography{emnlp2022}
\bibliographystyle{acl_natbib}

\clearpage
\appendix
\section{SUPERB Tasks Details} \label{sec:appendix_tasks}
We provide a brief overview of the tasks, prediction head settings and evaluation metrics. For further  details please refer to SUPERB paper~\cite{DBLP:conf/interspeech/YangCCLLLLSCLHT21} or the SUPERB leaderboard.\footnote{\url{https://superbbenchmark.org/leaderboard}}
\begin{itemize}
    \item ASR aims to transcribe audio into text. A vanilla 2-layer BLSTM is applied as the downstream task model, optimised with CTC loss. The evaluation metric is word error rate (WER).
    \item KS classifies utterances to detect specific keywords. Mean-pooling and a linear layer with cross-entropy loss are applied as the downstream task model. The task is evaluated using accuracy (ACC).
    \item QbE aims to detect spoken terms in an audio database by calculating whether a given query matches a spoken document. It does not require training. Dynamic Time Warping(DTW) and standard distance functions are used on all hidden states to report the final score. Maximum term weighted value (MTWV) is used for evaluation.
    \item SD, given an audio in which more than one person speak alternately, aims to determine the speaker at each timestamp. A single-layer 512-unit LSTM is applied as the downstream task model to SD task. The evaluation metric is diarisation error rate (DER).
    \item IC is designed to detect the intent of speaker from a spoken utterance. Mean-pooling and a linear transformation with cross-entropy loss are employed in the downstream task model. The evaluation metric is accuracy (ACC).
    \item SF aims to detect a sequence of semantic slot-types based on spoken words. Slot-type labels are included in transcriptions as special tokens, while SF is treated as an ASR problem. A vanilla 2-layer BLSTM with CTC loss is applied as the downstream task model. The evaluation metrics are slot-type F1 score and slot-value CER.
\end{itemize}

\begin{figure}[t]
    \centering
    \includegraphics[trim={6cm 3.3cm 4.5cm 3.5cm},clip, scale=0.11]{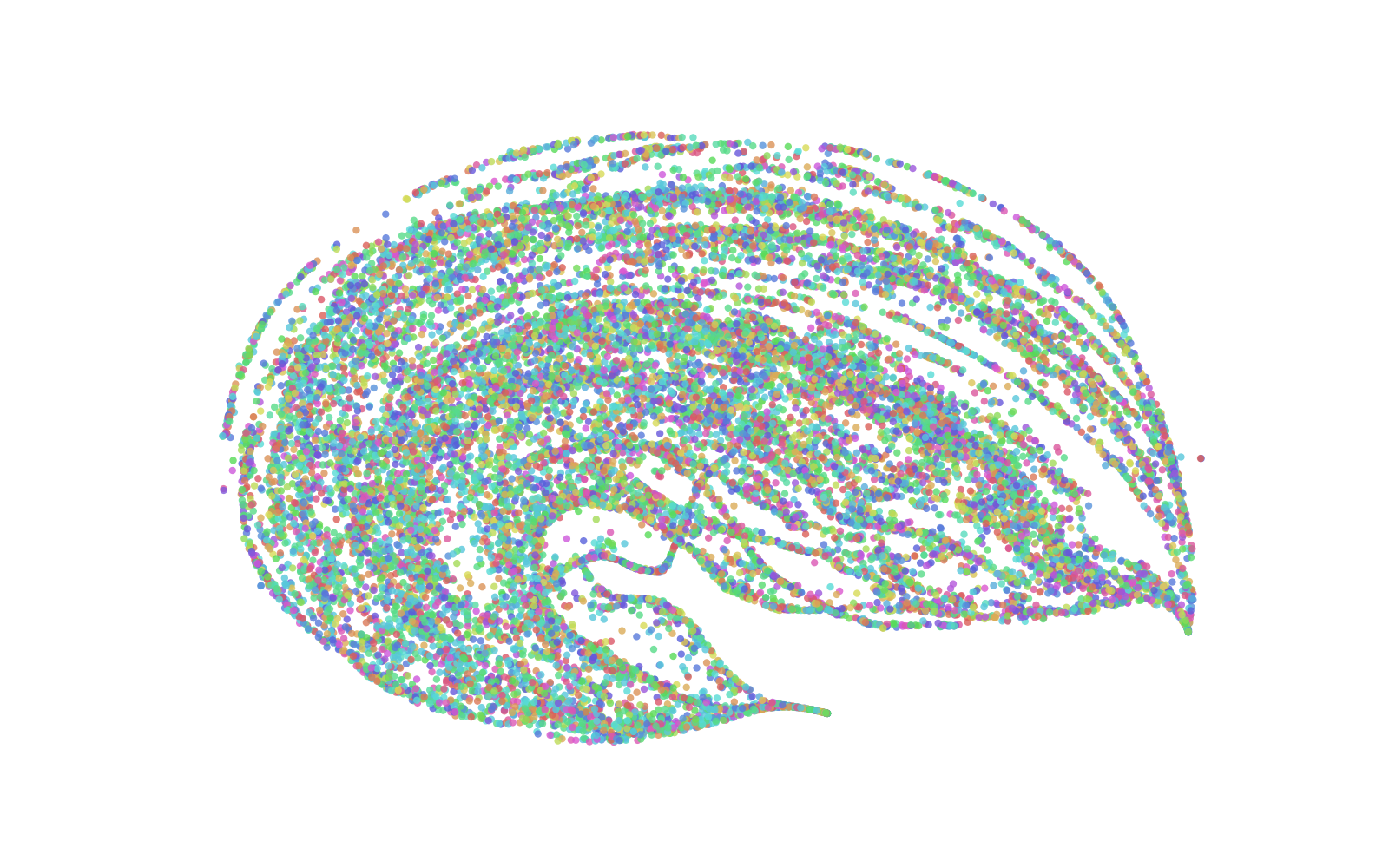}\hspace{3mm}
    \includegraphics[trim={6cm 3.3cm 4.5cm 3.5cm},clip, scale=0.11]{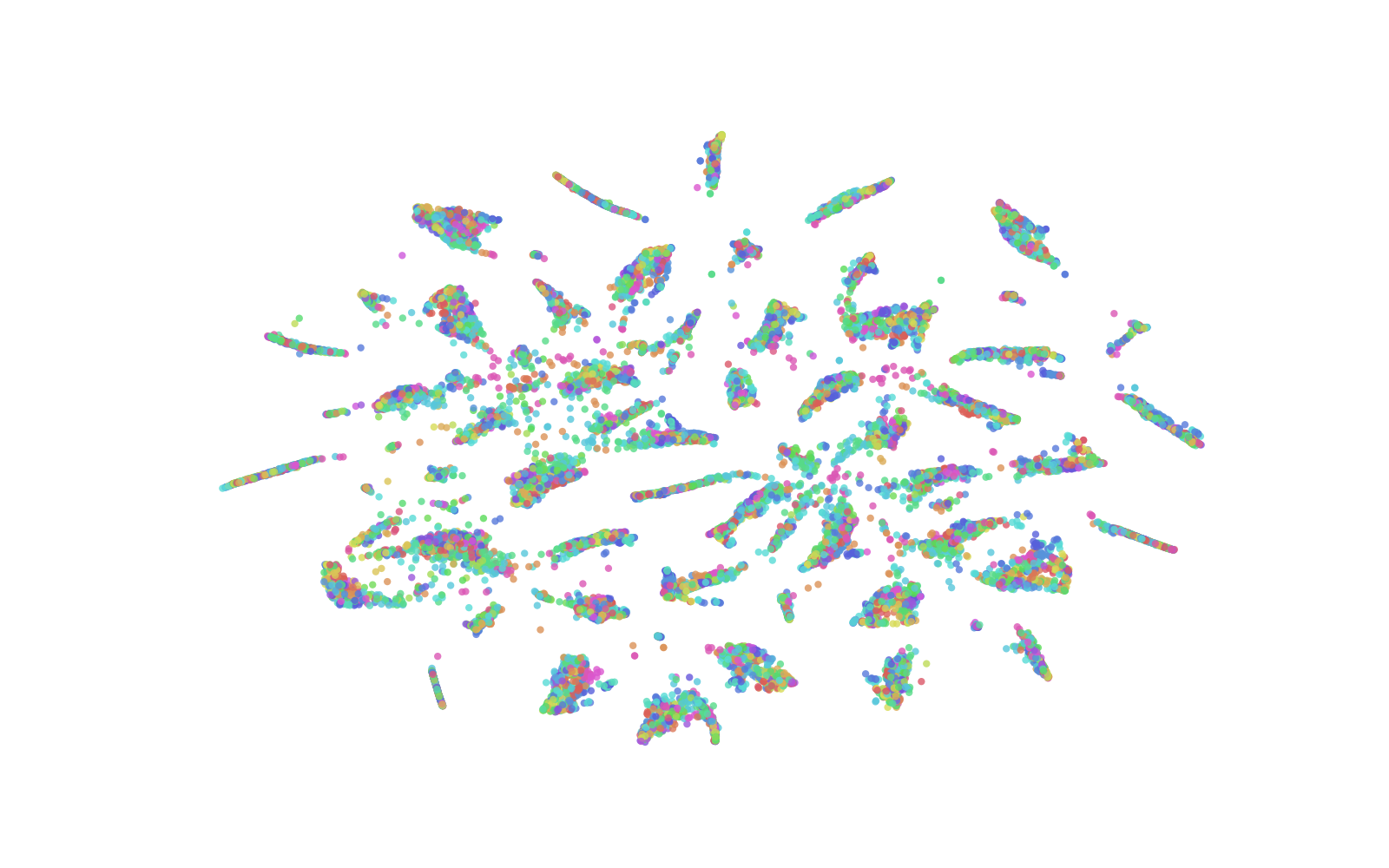}
    \includegraphics[trim={6cm 3.3cm 4.5cm 3.5cm},clip, scale=0.11]{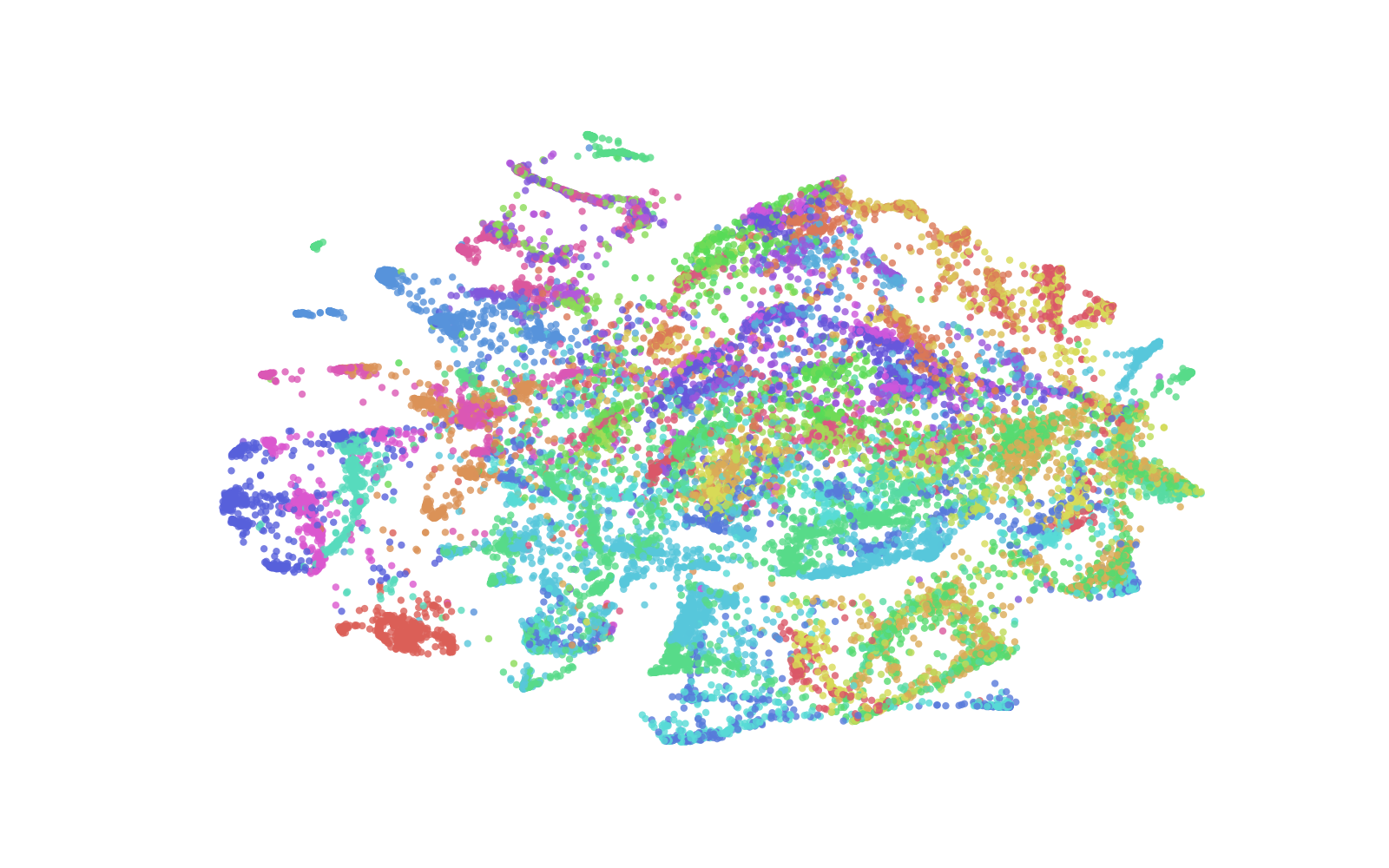}\hspace{3mm}
    \includegraphics[trim={6cm 3.3cm 4.5cm 3.5cm},clip, scale=0.11]{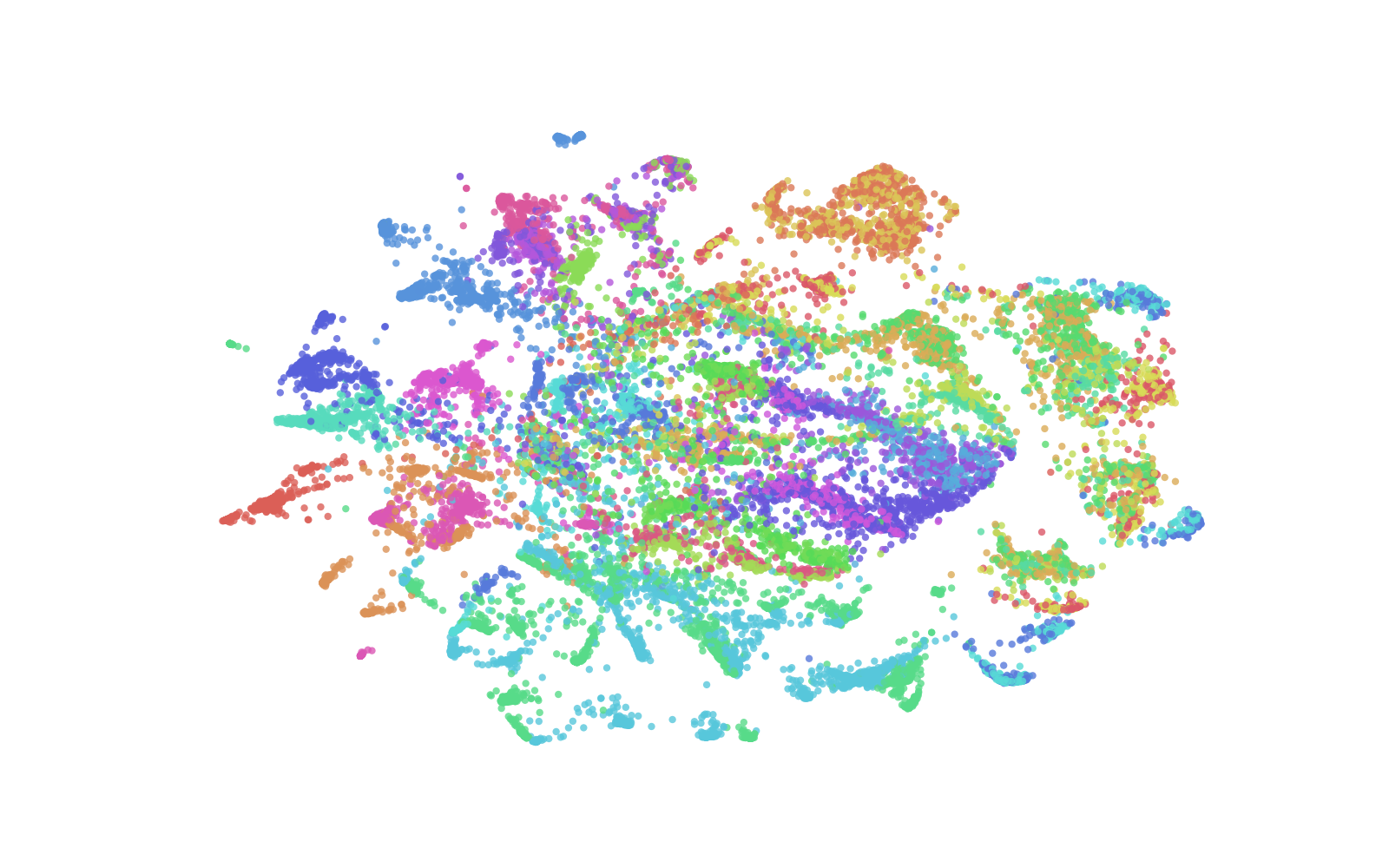}
    \caption{t-SNE visualisations of the representations from Intent Classification~\cite{DBLP:conf/interspeech/YangCCLLLLSCLHT21}  training set prior to fine-tuning. \textbf{Top-Left:} \textsc{W2V2}; \textbf{Top-Right:} \texttt{Twin}; \textbf{Bottom-Left:} \texttt{Neutral}; \textbf{Bottom-Right:} \texttt{Mixed}. Colours indicate class labels.}
    \label{fig:tsne-ic}
    \vspace{-0.3cm}
\end{figure}

\begin{table}[t]
\small
    \centering
    \begin{tabular}{l c c c }
        \toprule
        %  & & &\multicolumn{3}{c}{\# Max Step} \\
        %  \cmidrule(lr){5-8}
        & \multicolumn{3}{c}{ Training Instances} \\
        \cmidrule(lr){2-4} 
         \thead{Task} &1\%&5\%&10\%\\
         \cmidrule(lr){1-1}\cmidrule(lr){2-4} 
    SD &  20k & 20k & 50k      \\
    SF & 50k & 100k & 100k  \\
    IC & 20k & 20k & 50k \\
    KS & 20k & 50k & 50k\\
    % QbE & content  & 107.5k & 2438 || 138 || 138      \\
    ASR& 50k & 100k & 100k       \\
    \bottomrule
    \end{tabular}
    \caption{Maximum number of training steps.}
    \label{tab:max-step}
\end{table}

\section{Implementation Details}\label{appendix:training details}

We rewire \textsc{wav2vec 2} for 1.7k, 11.6k, 5.0k updates ($\approx$1, 7, 3 epochs) for \textsc{Twin}, \textsc{Neutral} and \textsc{Mixed}, respectively. {Applying more epochs may lead to overfitting, for \textsc{Twin}, training loss drops to almost zero in epoch 2.} Next, in downstream tasks, under the full-resource condition, we set the max step to $\sim$200k for both the baseline and our models. Under the 1\%, 5\% and 10\% resource settings, Table~\ref{tab:max-step} shows the max step we set for training models respectively, and test the best checkpoints accordingly. Note that the baseline and our models are trained with the same number of updates in all settings. 
During the rewiring process, we use dropout = 0.1 of \textsc{wav2vec 2} for all training instances. The learning rate is set to 1e-6 and the temperature for the infoNCE loss to 0.04. 
Additionally, it is perceived that contrastive learning requires a sufficient number of positive and negative pairs \cite{gao2021scaling}, which cannot be achieved naively, again, due to the length issue. To solve the problem, we use a batch size of 4. For each input within the batch, we get the augmented version and feed the two data points to the network to obtain two utterance-level vectors; this process applies to the rest of 3 training examples. This effectively relaxes the memory requirement for training pretrained speech models, given the hardware constraint. 
To avoid memory issues that come from lengthy audio signals, in practice we read one audio data at a time and set the audio length threshold to 90k. 
Whenever the audio length is greater than the threshold, we split the audio input into two parts with equal lengths, and randomly select one of them as the basis for further augmentation.  
Furthermore, to mask the augmented data in training \textsc{Twin}, we randomly pick a starting point in the first four-fifths of the frames of the audio data, and then mask consecutive frames that are one-fifth of its total length. For \textsc{Neutral}, we generated neutral speech from the transcriptions of the chosen LibriSpeech subset with the aforementioned TTS software offline. While training \textsc{Mixed}, we randomly select a augmentation technique from \textit{twin} and \textit{neutral} to construct positive pairs. The truncation trick applies to all three methods. 

\section{t-SNE on Intent Classification}
% \ehsan{Need to add a bit of text to explain Figure~\ref{fig:tsne-ic}}
Figure~\ref{fig:tsne-ic} illustrates t-SNE~\cite{JMLR:v9:vandermaaten08a} visualisation on IC. While the representation space is not as well-separated as KS, it is still rather clear that \textsc{Neutral} and \textsc{Mixed} have better separation of speech representations and consistency within clusters (higher concentration of same color within clusters). While the points seem to be separated across the space, a closer look indicates that the shaped clusters are substantially mixed, making it much more difficult for the task layer to discriminate between these points (much worse than \textsc{W2V2}). 
%The representations form lots of clusters which contain different class labels. This makes it more difficult for the downstream task model to discriminate the differences between representations with different class labels, which greatly hurt the performance. We will investigate the reasons for this phenomenon in future work.
\end{document}